# Active Observer Visual Problem-Solving Methods are Dynamically Hypothesized, Deployed and Tested

**Markus D. Solbach**            SOLBACH@EECS.YORKU.CA
**John K. Tsotsos**              TSOTSOS@EECS.YORKU.CA
Electrical Engineering and Computer Science, York University, Toronto, Canada

## Abstract

The STAR architecture was designed to test the value of the full Selective Tuning model of visual attention for complex real-world visuospatial tasks and behaviors. However, knowledge of how humans solve such tasks in 3D as active observers is lean. We thus devised a novel experimental setup and examined such behavior. The primary goal was one of discovery: we sought to explore and identify the characteristics of human behavior in active tasks that could then inform the development of cognitive systems such as STAR that perform similar tasks. We discovered that humans exhibit a variety of problem-solving strategies whose breadth and complexity are surprising and not easily handled by current methodologies. It is apparent that human solution methods are dynamically composed by hypothesizing sequences of actions, testing them, and if they fail, trying different ones. The importance of active observation is striking as is the lack of any learning effect. These results highlight the new dimensions of visuospatial problem-solving that active observers employ.

## 1. Visuospatial Tasks

The widely acknowledged success of modern computer vision coupled with the assertions that their methodologies and performance are human-like (e.g., He et al. 2015) may lead one to think that vision is a solved problem. The reality is far from this. The breadth of visuospatial tasks humans perform daily is stunning (Carroll 1996), and most have barely been considered by the computational community. One aspect of human vision that is insufficiently examined in computer vision as well as in cognitive architectures (CA) (review in Kotseruba & Tsotsos 2019) is visual attention. Of course, visual attention is widely acknowledged as important and many theoretical and practical investigations have appeared. However, they all seem limited as they include only the most obvious and basic of attentional mechanisms, rarely involving a fixation control component. Further, they are often accompanied by claims of biological correspondence or similarity (e.g., Bengio 2019), when in fact, there is a significant chasm between human attentional abilities and what current computer vision systems include (for background, see Itti, Rees & Tsotsos 2005, Tsotsos 2011, Carrasco 2011, Nobre & Kastner 2014, Moore & Zirnsak 2017, Goodhew 2020).
    The STAR (Selective Tuning Attention Reference) architecture is based on this assertion: if the Selective Tuning (ST) model of attention actually represents the set of brain mechanisms of attention, then ST should provide all the attentional support a behaving brain (agent) would require. ST includes a broad and comprehensive set of mechanisms, many of them originally unintuitive

but now with significant experimental support.[1] In order to test the assertion, the full set of other functional components, as also found in most other CAs, must be connected to ST (Tsotsos & Kruijne 2014). This set includes fixation control in 3D space, something not common among CA's nor computer vision. STAR also does not assume that observers are passive, that is, that they are simply receivers of externally determined input and play no role in selecting what, how, why, when and how to observe their external world (Bajcsy et al. 2018). However, the study of human visual behavior as an active observer has been limited. As a result, any potential testbed for STAR seems unavailable, as are even particular exemplar problems with which to test our ideas. The comprehensive description of human visuospatial abilities in Chapter 8 of Carroll (1996) proved to be very helpful. We chose as a first exemplar the *same-different* task, a task humans need to solve often, and which seems a basic component of many other tasks.

Deciding if two objects are the same or different may seem straightforward. Often, we design objects to be easily discriminable, say by colour or size or pattern, but this is not always the case. Consider a task where you are given a part during an assembly task and need to go to a bin of parts in order to find another one of the same (e.g., assembly of furniture). Playing with interlocking toy blocks requires one to perform such tasks many times while constructing a block configuration, either copying from a plan, mimicking an existing one or building from one's imagination. There are many more examples. Obvious instances of this problem are not effective as probes into human solutions because humans are remarkable in their ability to home in on a workable strategy that can be used for most instances. We thus needed to push an experimental design to the extreme in order to discover the characteristics and limitations of the human solution space. The key question remains: What is the sequence of visual actions to correctly determine if two objects are the same? This problem has equal interest for human behavior as well as robot behavior.

In the current AI and computer vision community, one's first approach might be to learn solutions. It is quite likely possible to learn a viewing policy that simply covers all parts of a viewing sphere around each object, and then compares the feature representations on the sphere surface. There are an increasing number of works that attempt to actively learn visuospatial tasks, incorporating prior knowledge to varying degrees and using multiple observations (e.g., Settles 1995, Ren et al. 2020). But obvious, brute force solutions do not illuminate how humans do this in a far more efficient manner – we solve simple cases quickly, take an increasing number of views with increasing task difficulty, rarely need to see a full spherical view, and almost never take a single complete set of views. However, such conclusions seem subjective, and no experimental evidence is available that explicates how humans solve such tasks. Related visuospatial tasks have also been studied by the cognitive neuroscience community and as shown in Tsotsos et al. (2021), they seem to be converging on the importance of flexible and dynamic composition of processing elements to achieve solutions for a given task. This is consistent with Tsotsos (2011), namely that, vision seems to involve a general-purpose processor that can be dynamically tuned to the task and input at hand, and attention is a set of mechanisms that tune and control the search processes inherent in perception and cognition. However, dynamic tuning cannot occur without an explicit executive controller, as argued in Tsotsos et al. (2021). To test this view and confirm consistency with the neuroscience view of flexible composition, data is required, but little is available. The experiment described here may be helpful.

---

[1] The ideas behind ST appeared in Tsotsos (1988, 1990), and Tsotsos et al. (1995). For full description see Tsotsos (2011).

The remainder of this paper proceeds as follows. The next section begins with a brief description of the general idea of the experiment, followed by a summary of the experimental environment and setup, the stimulus set, and a small sampling of our results and observations. Section 3 introduces *Cognitive Programs* that promise sufficient power and flexibility for the representation of active observer behavior. A summary of results concludes the paper.

## 2. Examining the Same-Different Task

The classic instance of the same-different task is widely known from the work of Shephard & Metzler (1971). There, they used objects formed by concatenations of cubes, and depicted as black line perspective drawings on a white background. Subjects were shown pairs of these objects and were asked if the objects were the same or different. Stimuli were 4-5cm in linear extent, seen in two windows, viewed from 60cm. In other words, subjects were passive viewers with a constant target visual angle for each stimulus object. The "view" was pre-determined. Since reaction times were as long as 5s, there was plenty of time for eye movements, but no report of them was provided. Results showed that subjects mentally rotated one object into the other, this being an inference made by considering response time. However, something important is missing here. Humans did not evolve categorizing 2D line drawings on a screen; thus, we sought to move this classic study into a more ecologically valid setting. We push this experiment to its limits, increase stimulus complexity, perform it in three dimensions, allowing observers to choose whatever viewpoints they wished towards solving the task, and record in detail how they viewed the objects. In this way, we wished to discover exactly how a human agent solves such a real 3D visuospatial task. First, we needed to decide on the object set.

### 2.1 Test Objects

Our goal was to create unfamiliar objects of the same sort as those of Shepard & Metzler, but in 3D and with controlled and measurable complexity. We began with a simple base (seen in the top panel of Figure 1, bottom row, item 1), to which we then added simple blocks, such as seen in the bottom row of the top panel, labelled 2 and onward. The objects then evolved by repetitions of this action. The complete object set is labelled $L_1$ and its subset, $L_2$, seen in Figure 1[2]. These objects, other than the simplest, may not seem to satisfy our ecological validity goal. However, we need to both satisfy that goal as well as push the visual system in order to uncover its limits. These objects are the stimuli placed within the experimental setup described in the next section.

### 2.2 Psychophysical Experimental Setup for Active Observers

Most past and present research in vision, both computer and human, involves passively observed data. Humans, however, are active observers in their everyday lives; they explore, search for, and select what and how to look (Bajcsy 1985, 1988; Bajcsy et al. 2018). Nonetheless, how exactly active observation occurs in humans is an open problem. We developed Psychophysical Experimental Setup for Active Observers in a 3D world (PESAO), a unique experimental facility.

---

[2] For more detail, plus a link to the 3D printable models see Solbach & Tsotsos (2021).

The goal was to design PESAO for various active perception tasks with human subjects as active observers. While many psychophysical studies explore human performance, there seem to be none that involve active observers with 3D real stimuli. PESAO allows us to bring many studies to the 3D world and to involve active observers. The custom hardware and software combine precise head motion tracking in full 6DOF with gaze tracking at microsecond resolution while the subject is entirely untethered and can move freely and naturally. In sum, the system provides tracking and recording of eye fixation, first-person video, head-mounted IMU sensor, bird's-eye video, and experimenter notes. Head-tracking accuracy is 0.4mm in 97% of the capture volume, while Gaze-tracking accuracy is 1.42 degrees error in visual angle. More detail is in (Solbach & Tsotsos 2020, 2021).

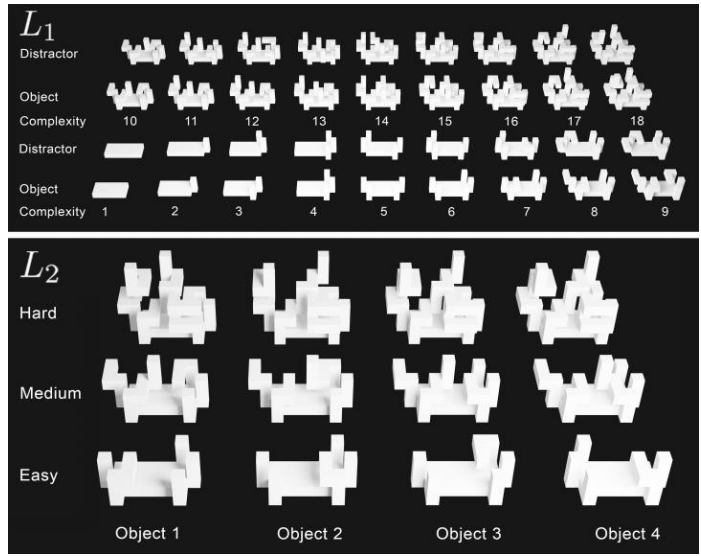

*Figure 1.* Top: Illustration of $L_1$ with all 36 objects. Bottom: Illustration of its subset, $L_2$ with 12 objects, split into three different complexity levels. More detail may be found in Solbach & Tsotsos (2021).

### 2.3 Experiment

For the objects in our first experiment, we choose the $L_2$ set shown in Figure 1. This set offers three complexity classes with four different objects each. Our definition of *same* and *different* is entirely based on the geometrical configuration of the objects. *Same* means that both objects consist of the exact same configuration of elements (including size), whereas *different* means that at least one element of the geometrical composition of at least one object is different. In order to analyze the active behaviours of observers, we track the head motion and the participants' gaze, and allow natural movements with subjects untethered and without predetermined viewpoints.

All variables, except object complexity, are chosen randomly using a uniform distribution for each trial. For object complexity, we opted for a pre-defined set of six trials of each complexity level; however, the order of presentation was again random using a uniform distribution. No pattern due to biases in the sampling was observed.

To date, we have conducted 846 trials of the 3D same-different task with 47 randomly sampled participants. Two objects are selected and mounted on posts in the experimental space. There are multiple independent variables to explore: objects may be *same* or *different*, object complexity may be *easy*, *medium* or *hard* (as shown in the $L_2$ set of Figure 1), the object pose or mounting orientation difference between their base plates may be 0.0°, 90.0° or 180.0°, and the subject's starting position may be at the *long* (close to targets, each equidistant from subject), *corner* (subject oblique to targets, one closer than the other), or *short* positions (targets are in line and subject views along that line, one object farther than the other). All are chosen randomly for each trial and subjects are placed into the experimental space facing away from the objects at one of the 3 starting positions. Once the subject is instructed to turn towards the objects, the trial begins, and all head and gaze movements are recorded, as are the first-person and third-person videos of the entire trial. The first fixation that counts is the one that falls on a target object. After the trial, subjects complete a questionnaire that asks about strategies and observation methods. Figure 2 shows a sample gaze trace for a particular trial.

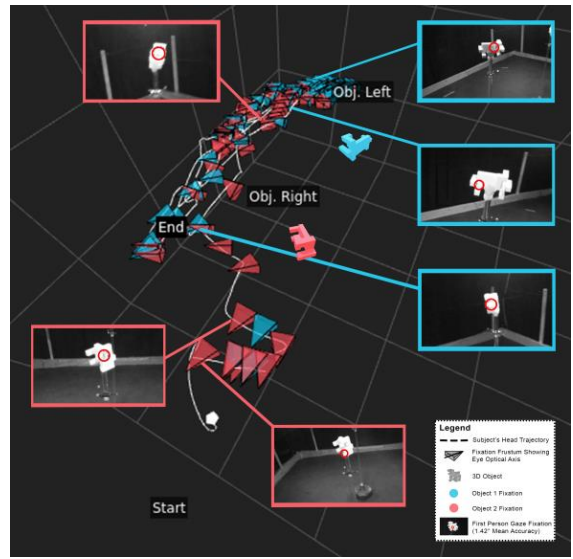

*Figure 2.* A visualization of the recorded data using PESAO. The subject's movement is plotted as a dashed line in white, and fixations of either object are illustrated as a fixation frustum in the corresponding colour of the fixated object. Selected fixation frusta are annotated with snapshots of the subject's first-person view and the gaze at a particular fixation (red circle). In this example, the objects are the *same*, they differ in pose by 90°, and the subject started from the *short* position.

There was no hypothesis that these experiments were testing and there was no learning task for the subjects. As mentioned, this was an experiment of discovery. We sought to explore and identify the characteristics of human behavior in active tasks because little seems known about this. What we find could then inform the development of artificial systems that perform similar tasks.

**2.4 Some Results**

We considered several different performance metrics such as accuracy, response time, the number of fixations and more. All of these were computed with respect to the different variables used in this study: starting position, object complexity, orientational difference, and target sameness. Furthermore, we also investigated how a correct or incorrect trial affects these measures. Only a sampling of the main results will be given here; the full report is Solbach & Tsotsos (forthcoming). The section titles here refer to the three variable dimensions examined in each data plot, in *ordinate-absicca-applicate* order, even though plotted in 2D only (applicate described in the figure legend).

*2.4.1 Accuracy - Object Complexity - Subject Starting Position*
Humans are remarkably good at this task (Figure 3A). Throughout all evaluated combinations of this task, participants achieved an absolute mean accuracy of 93.82%, σ = 3.9%. The best performance was achieved for the complexity level *easy*, with subjects starting by viewing from the *long* position with an orientation difference of 0.0°. These objects are the least complex, making them the easiest to compare. Secondly, starting from the *long* position presents both objects at the same distance and next to each other immediately from the start. An orientation difference of 0.0° means that the objects are also aligned in their orientation, simplifying the comparison. Further, for all complexity classes, trials starting from the *long* position always resulted in the best performance within a complexity class: 96.96% (*easy*), 94.52% (*medium*), and 96.2% (*hard*). By contrast, objects of complexity *hard* for trials that started from the *short* position have the worst mean performance of 90.38%.

*2.4.2 Number of Fixations - Object Complexity - Task Ground Truth (same or different)*
Across all trials, the absolute mean number of fixations was 92.38 (see Figure 3B). No trial took fewer than 6 fixations[3]. In terms of trends across the different aspects of the experiment, it was seen that *same* response required more fixations (roughly 10 - 20 more) than *different* responses. In general, harder instances took more fixations, error responses took more fixations, greater orientation differences required more fixations for some configurations, but not all, to reach a conclusion. The *long* starting position seemed to require the fewest fixations, the *short* position required the most for *easy* and *hard* objects, while the *corner* position required the most for *medium* objects.

*2.4.4 Response Time - Object Complexity - Subject Starting Position*
Response time is the time elapsed from the start of the trial (the subject turns around and faces the objects) to the end of the trial (see Figure 3C). On average, the response time over all trials was 47.52s, with σ=30.39 from start to answer. Among all our trials, the shortest response time is for *medium* cases starting from the *short* position (4.2s), whereas the longest response time is 298s for a *hard* case from the *long* position. While incorrect answers are few, it can be noted that the response time, in general, is higher than for a correct answer. While the difference for the *easy* case is about 20s, the difference is less significant for the *medium* case with around 9s and roughly 11s for the *hard* case. This shows that uncertainty about the answer prolongs the response time and results in more extended observations. While we can see a drop in response time of about 5 seconds from complexity level *easy* to *medium* (37.5s and 32.5s, respectively), the response time increases significantly to 49.5s for the *hard* cases. We conclude that though the objects are roughly of same size, the increase in visual features demands more time.

---

[3] There is an important point here. Even our simplest case required 6 fixations, meaning 6 feedforward passes through the visual system, with possible top-down components for each, in addition to the computation and execution of eye fixation itself. Models of vision that employ a single feedforward processing pass with no eye movements seem unsuited to this task or to comparison with human vision.

*2.4.6 Accuracy - Trial Number - Object Complexity; Fixations - Trial Number - Object Complexity*
We compared accuracy and fixation numbers for individual subjects across the number of trials each subject performed (see Figures 3D, E). We have 47 subjects, each subject did 18 trials (6 for each complexity level), and no target object configurations were repeated. Nevertheless, we expected that some improvement in performance or reduction in the number of fixations would begin to appear. This was not the case; there is no discernable learning effect in accuracy, number of fixations or overall head movement. No learning effect can be seen for the case of *easy* complexity. However, a minimal effect can be seen for the complexity *medium* and *hard*. For the *medium* cases, the first trial took, on average, about 80 fixations, and the last trials took, on average, about 50 fixations. Similarly, for the *hard* cases, a decline in fixations can be seen; about 150 fixations at trial 1 and about 100 at trial 6. However, the minimum for the *hard* cases was actually reached at trial 5 with about 75 fixations.

*2.4.3 Head Movement - Object Complexity - Relative Target Orientation*
The absolute mean of head movement was 16.62m over a trial and no trial had less than 1m of head movement (see Figure 3F). Error responses were accompanied by significantly more head movement and head movement generally increased as problem instances became more difficult. The amount of head movement slightly increased from complexity cases of *easy* to *medium*, but increased more distinctly for *hard* cases. A clear trend can be observed between the amount of head movement and amount of orientational difference; if both objects were aligned (0°) in all complexity cases, the least amount of movement was required, for 90° an increase of 2-5m on average was recorded, and finally, for the 180° an additional increase of 1-5m was recorded.

*2.4.5 Accuracy - Object Complexity - Relative Target Orientation*
We also evaluated the effect of the relative orientation of the objects (see Figure 3G). For the *easy* case, there is a clear gradient of accuracy following the increase of orientation difference. Notably, trials of *easy* with orientation 0.0° had an accuracy of 100%. However, for the other complexity levels, a different pattern can be identified; 90° was most accurately identified with 94.82% and 90% for *medium* and *hard*, respectively. 0° and 180° ranked second and third. Orientation difference also seems to impact the number of gaze shifts and head movements required. The amount of head movement and response time increased with orientation difference at all complexity levels, but there was no clear conclusion regarding accuracy.

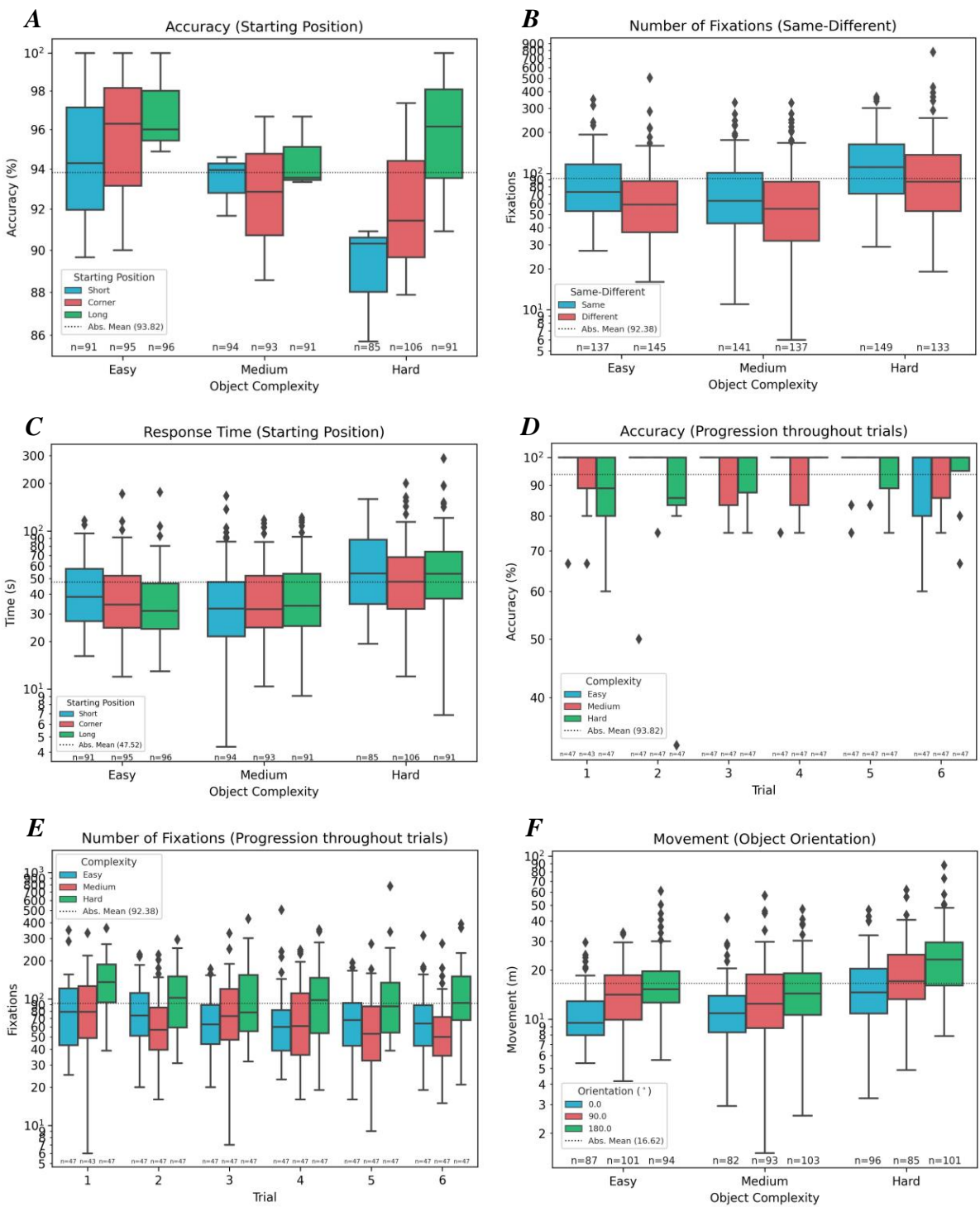

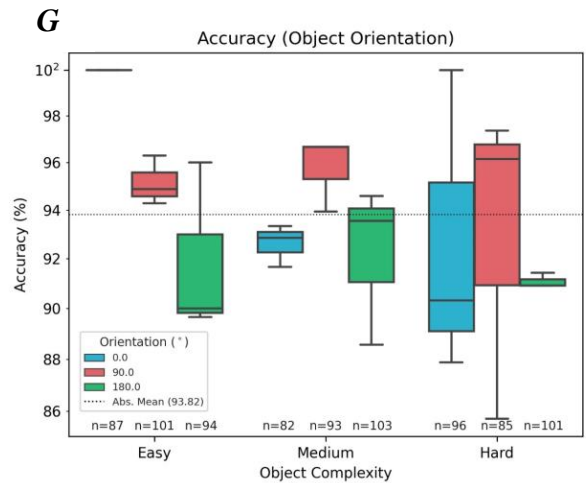

*Figure 3.* Some experimental results for the Same-Different task. All plots use standard box-and-whisker format. Below each box, the number of subjects for that box is given (n=). The third variable (applicate) is described in the legend for each plot. More detail can be found in the text.
**A:** *Accuracy - Object Complexity - Subject Starting Position.* Overall, accuracy decreases with the increase of complexity. The *long* position start always resulted in the best accuracy. Except for the *medium* complexity case, the sequence of increasing accuracy was *short*, *corner*, and *long*. For the *medium* case, however, this order was *corner*, *short*, *long*. Absolute mean accuracy was 93.82%, with σ=3.9%.
**B:** *Number of Fixations - Object Complexity - Task Ground Truth (same or different).* For all complexity levels, *different* objects resulted in fewer fixations on average. While trials of *medium* complexity needed the least fixations, trials of *hard* complexity needed the most. An outlier *hard* trial was seen with almost 900 fixations. For trials with *medium* complexity, which were *different*, as few as 6 fixations were used.
**C:** *Response Time - Object Complexity - Subject Starting Position.* The absolute mean was 47.52 seconds for a trial (start to answer) with σ=30.39s. *Hard* complexity trials took longer than *easy* complexity, and *medium* complexity trials were answered the fastest on average. A wide variation in response time is noted; the fastest response time was recorded for *medium* complexity starting from the *short* position (4.2s), and the longest response time was recorded for *hard* complexity starting from the *long* position (298s).
**D:** *Accuracy - Trial Number - Object Complexity.* Every subject performed 18 trials (6 each of *easy*, *medium*, *hard*). However, no discernable learning effect was detected. The plot presents the average accuracy for each of the six trials for all three complexity classes. Surprisingly, for *easy* complexity, trial six was the worst. This is counterintuitive as one could think that accuracy will increase with practice.
**E:** *Fixations - Trial Number - Object Complexity.* No learning effect on the number of fixations can be seen for the case of *easy* complexity, with minimal effects for the other cases of *medium* and *hard*.
**F.** *Head Movement - Object Complexity - Relative Target Orientation.* The absolute mean was 16.62m of head movement from start to end of a trial. The amount of head movement slightly increased from complexity cases of *easy* to *medium*, but increased more distinctly for *hard* cases.
**G.** *Accuracy - Object Complexity - Relative Target Orientation.* For the *easy* case, 0° difference (aligned objects) resulted in 100% accuracy (overall trials). 90° ranked second for the *easy* case with 94.5%, and 180° resulted in the lowest accuracy with 90%. Interestingly, while one might suggest that an increase in orientational difference also results in a decrease in accuracy, this is not entirely true. While 180° resulted in the lowest accuracy across all complexity classes, it is actually 90° which resulted in the best average accuracy for the *medium* and *hard* cases.

Although we continue to collect trials, this early result on lack of learning effect was most surprising. It may point to the possibility that this task is not 'learnable' by humans, that their high level of performance is not dependent on what they learn from previous attempts or exposures but rather is due to their on-the-spot strategy composition for the particular task scenario of the moment. This connects well with the conclusion of the review in Tsotsos et al. (2021), where the importance of flexible composition of elements to achieve solutions for dynamic task presentation was noted. Subjects might of course learn how to better create new compositions, but we have not yet analyzed this meta-level aspect.

**2.5 Mining the Fixation Sequences**

As the data presented above imply, the goal of finding structure within sequences of hundreds of actions encompassing about 80,000 fixations is a daunting task. Most data-mining approaches were not helpful; even the popular Trajectory Pattern Mining Algorithm (Giannotti et al. 2007) did not bear fruit, finding either nothing or large meaningless patterns. We examined the subject questionnaires for clues. For example, a common report was that subjects look at one object then the other, searching for the presence of particular structures seen in one object on the other. This and other such reports provided a few abstract anchors and using these, we embarked on hand-labelling the data. We looked through the first-person videos, annotated the gaze and looked for commonalities. When found, each was formalized and checked across other trials.

This work continues but at present the high-level abstract patterns seen are the following:

A. The *Initialization* stage is usually a short time window between 3-5s in which the subject performs 2 routines to set themselves up for the development and deployment of a strategy.
   1. *3D Layout* - Subjects seem to first evaluate the overall 3D layout. After receiving the start signal, the subject turns away from the black curtain and assesses the scene for traverseability and to self-localize within the scene. This is detectable as a quick scan of the environment, such as panning the field of view from one side of the experimental setup to the other.
   2. *Location of Target Objects* - The second routine deals with localizing the objects within the scene. The subject needs to know where the objects are and their spatial relationship in order to plan the first spatial operation. This operation is often intertwined with *3D Layout*; distinct, short fixations (< 300ms) of both objects, either during or after *3D Layout*.
B. Subjects seem to think about strategy. The *Strategy Formulation* stage is the core stage, which might be repeated several times by developing, deploying and dismissing different strategies (hypothesize-and-test). A strategy is comprised of one or more, with possible recurrence, of:
   1. *Global Gist* - This describes how much of the virtual viewing sphere around each object has been explored. We quantize the viewing sphere into eight sectors. A sector is considered visited if fixation is carried out from within it. While the head pose needs to be within this sector, the gaze data is used to determine whether the object is fixated.
   2. *Outlier Detection* - An outlier is an element of one object that differentiates from the other object and is determined using only the gaze information. If the subject observes the differentiating part of the object which was previously observed on the other object and finalizes the trial with an answer, *outlier detection* is noted.
   3. *Divide and Conquer* - This operation divides the object into smaller parts which are used for comparison. If the part has been divided and compared with the other object, this part is considered 'conquered.' To detect this operation, the head pose, as well as the gaze, is used

      to recognize which part is being sub-divided and compared. A conquered part is not further sub-divided but can be revisited.

      4. *Coarse to Fine* - Similar to *Divide and Conquer*, this operation reduces the visual information, but in this case, in time. This operation is characterized as increasing the amount and duration of fixations on parts of the object.

      5. *Alternating Fixation* - This emphasizes comparative fixations of both objects. It is characterized as repeatedly fixating between the same parts of each object at least twice while the head mainly being stationary.

      6. *Alternating View* - This operation is similar to *Alternating Fixation*, however, it also includes a change of viewpoint (head movement).

C. *Confirmation* - This stage acts as the control module to verify whether a potential strategy provides the correct answer.

  1. *Strategy Repetition* - Interestingly, once a strategy has led to an answer, the subject sometimes repeats the strategy at least once. In other words, double or triple-checking the answer. If during the confirmation stage uncertainty arose, the subject enters *Strategy Formulation* to refine or dismiss the strategy.

D. There are a number of specific visual and spatial operations employed in the above strategies.

  Visual Operations describe the types of gaze movements and seem to be of two kinds.

      1. *Tracing Connected Components* - This elemental operation is defined as a series of fixations spatially close together, e.g., following the object's silhouette.

      2. *Comparing Arbitrary Components* - Instead of following a geometry with multiple fixations in close proximity, this elemental operation is characterized by more considerable distances between fixations, e.g., observing faces of the object punctually.

  Spatial Operations describe the physical movement of the sensory apparatus from one spatial location to another and are also of 2 kinds.

      1. *Point of View Change* - It is defined as moving the entire body to a different spatial location beyond the motion of the sensory apparatus (i.e., eyes).

      2. *Viewing Angle Change* - Defined as moving the sensory apparatus only.

The italicized phrases just defined are the ones used in the directed graphs, shown in Figures 4A, 5 and 6, extracted for each trial and which summarize the full sequence of actions a subject exhibits. Previously, in Figure 2, a set of viewing directions (frusta) was shown without most of the temporal links or annotations nor functional interpretations. Figure 4A is a 'bird's-eye-view' figure (but can be downloaded and examined on a large screen as the caption details) showing a single trial in order to give an impression of the scale of the solution space. This trial is what a subject did for the initial conditions of a pair of objects from the *easy* group of Figure 1, with an orientation difference of 90°, the objects in fact being the *same*, and the subject starting from the *long* position. The point of showing it in this form is to emphasize global characteristics. A portion (from t=3.40s and onward) appears in more readable form in Figure 5.

    The first thing to point out is that subjects do not take a direct path to a solution. The observed sequences almost always included several apparent trial and error components. One might wonder about these results since the target objects seem complex and perhaps not so realistic. However, in early pilot tests, the use of simpler structures (for example, the ones in the bottom two rows of the

$L_1$ panel of Figure 1), we observed that subjects quickly found a solution strategy that always led to the correct answer - count the number of blocks. To do so still required a number of fixation changes and a number of body and head movements, but the task was thus too easy and not instructive. We needed to push the limits of the task.

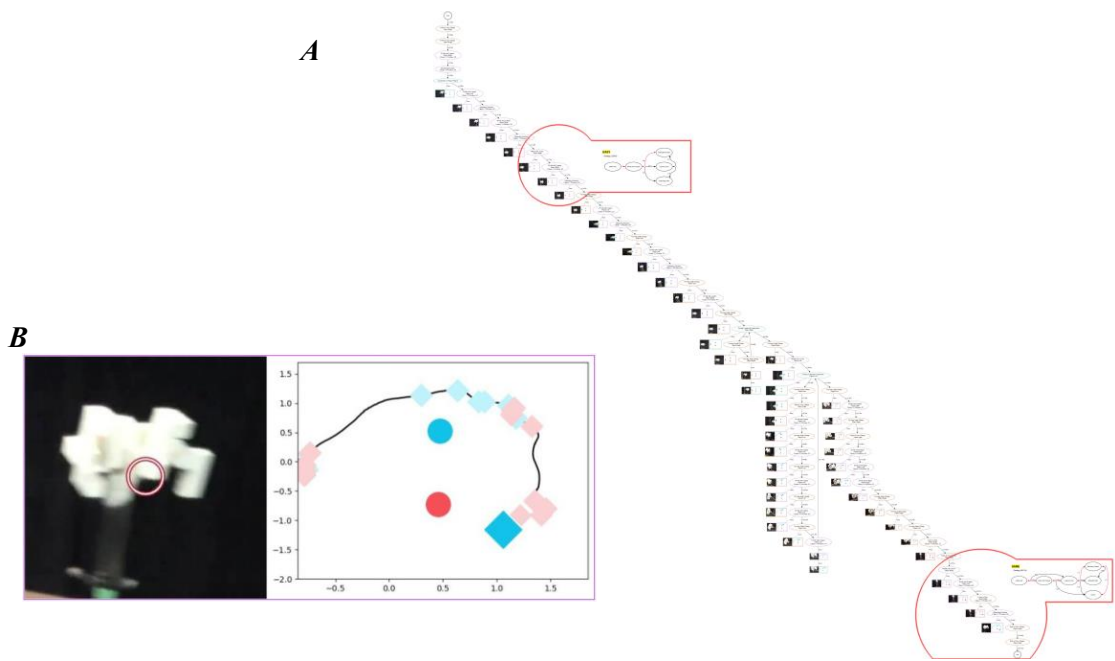

*Figure 4*. **A:** This 'bird's-eye view' figure represents the sequence of actions taken by a subject for a pair of target objects from the *easy* group, with an orientation difference of 90°, the objects being the *same*, and the subject starting from the *long* position. This was the 9th trial this particular subject completed. The trial required 24.19s and the subject performed 32 fixation changes with a total head movement of 10.05m. The final response was correct. The beginning of the trial is at the top. Two particular sub-graphs are highlighted with red circles (see Section 3.2 for more). The actual strategy identified is repeated in the right side of the red outline. Detail for the upper one can be found in Figure 5. **B:** Each of the dark blobs part A of this figure represents a particular gaze as shown here. On the left is the actual first-person camera-view with the red circle showing the point of gaze. The right portion shows the two target objects (red and cyan circles), the subject position (red diamond if viewing the red object, cyan otherwise) and the path traversed by the subject from the beginning to the current fixation. Smaller diamonds along the trajectory path indicate past fixations. The progression of this path is easily seen once the graph is magnified. The rejected solution hypotheses are seen as downward threads of actions not leading to an answer. The full resolution figure, best viewed with high magnification, is available at
https://data.nvision.eecs.yorku.ca/active/action_seq.jpg

A brief consideration of the scale of this problem in terms of size of its potential solution space is instructive. As described, we observed an average of 93 fixations during an average of 48s; this may encompass significant non-visual activity. With 300ms per fixation change, this leaves over 20 seconds for 'acting' (moving the head or walking) and 'thinking' (e.g., reasoning, planning, decision-making, working memory). The number of possible physical states of the experiment alone is enormous. Each fixation is defined by head pose (6DOF), fixation angle (the angle away from fully straight-ahead given head pose), and fixation depth (convergent binocular vision). The normal human visual field at the retina is 180° x 100°. At 2° x 2° quanta (measurement error of PESAO), there are 4500 possible fixation angles. Similarly, the head may rotate left-right about a stationary body approximately another 180 degrees and up-down another 180°. Say this is divided into 5° quanta, giving us 1296 possible head poses. Assume that the subject's body position may be located anywhere within the 4.3m x 3.4m PESAO space; let's quantize this space into 0.4m x 0.4m units for convenience, so there are roughly 91 possible positions. A subject's body may be oriented in any direction across 360° rotation, quantized in 5° units, this means roughly 72 possibilities. The physical states of the subject are the product of these numbers, over $3.8 \times 10^{10}$. With respect to the target space, the $L_2$ set has 12 objects, from which we choose 2 for each trial, and they may be the same. The choice of two is always from the same complexity subset and there are three subsets. Subject starting position matters as it affects the spatial order of presentation. Add in 3 relative pose orientations, and we arrive a total of 378 different target starting configurations[4]. The space of sequences of subject and target configurations, to cover the 6 to 800 fixations we have observed, is more difficult to estimate but clearly is an unmanageably large number.

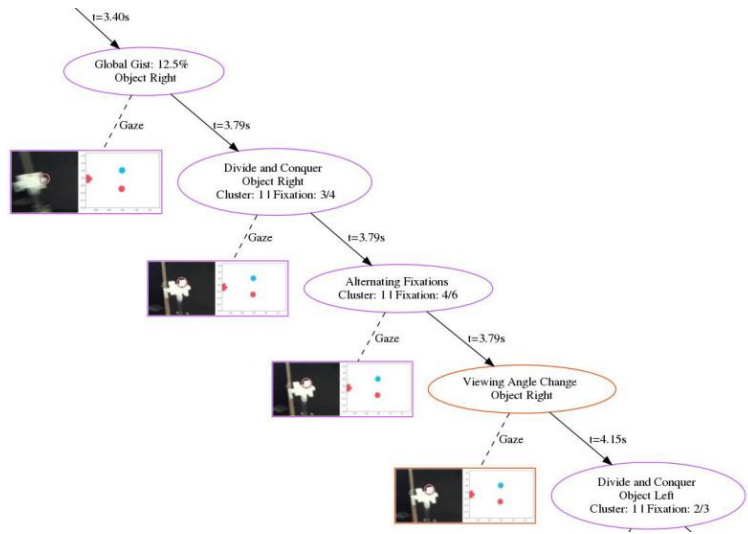

*Figure 5*. A zoomed portion of Figure 4A beginning at the t=3.40s mark of the trial. As can be seen, most of the sub-elements include pictorial annotations of the actual action taken by the subject at that time as well as the actual point of gaze for that observation. The top three ovals comprise one path through the upper strategy outlined in Figure 4A, further described in the top portion of Figure 6.

---

[4] For the *short* and *corner* starting positions, spatial positioning of targets matters so there are 16 possible pairs of the 12 objects; for the *long* position, because target objects are equidistant, there are only 10. The number of object configurations is computed for these two start positions separately and then summed.

## 3. Cognitive Programs

As the experimental results presented in Section 2.4 and the examples in the above figures show, the sequence of observations humans use to solve the same-different task can be long and complex, perhaps unexpectedly so. That they seem dynamically composed, deployed and evaluated, dependent on the task type, complexity and initial conditions, with little learning effect, seems counterintuitive, especially when compared to modern computational attempts at active learning (however, see Taylor et al. 2021 for review and a promising change). For the simple cases, our subjects showed simple, more routine methods, and for complex tasks, more involved strategies. Human intelligence is not designed for the simple case, but rather so that it can solve the most difficult as well. Simple solutions do not easily generalize, and our experimental design served us well in that it unveiled those situations.

Thus, a series of important questions arise: how do observers arrive at these strategies? Is the sequence determined all in advance, or is it constructed in parts, each depending on the result of the previous? How can these sequences be represented? Are portions of them learned in past experience and then used as needed? These human sequences are many seconds long, and thus the processing is clearly not all confined to the visual cortex, where one feedforward pass requires only about 150ms. What might these extra-visual actions be, how could they be inferred from the observed sequences, and how can they be expressed in algorithmic form? Early steps towards answers follow.

### 3.1 Cognitive Program Concept

Cognitive Programs (CPs) are a modernized offspring of Ullman's seminal Visual Routines (Ullman 1984) and provide an algorithmic structure to visual, cognitive, action and attentional behaviors. Ullman's proposal addressed how human vision extracts shape and spatial relations. He proposed that: visual routines (VRs) compute spatial properties and relations from base representations to produce incremental representations; VRs are assembled from elemental operations; new routines can be assembled to meet processing goals; universal routines operate in the absence of prior knowledge, whereas other routines operate with prior knowledge; mechanisms are required for sequencing elemental operations and selecting the locations at which VRs are applied; and, VR's can be applied to both base and incremental representations. VRs have been studied and supported by both computer scientists and brain scientists and seem an enduring idea (see review in Tsotsos & Kruijne 2014). However, the idea is dated in its detail and requires updating. Cognitive Programs (Tsotsos, 2010; Tsotsos & Kruijne, 2014) are an elaboration and modernization of the original idea and are based on a broader up-to-date view of visual attention and visual information processing. CPs are sequences of representational transformations, attentional tunings, decisions, actions, and communications required to take an input stimulus and transform it into a representation of objects, events or features that are in the right form to enable the solution of a behavioral task. *Methods* are like Ullman's universal routines, while when parameterized for the current situation, they are termed *Scripts*. In a real sense, CPs are algorithms for solving problems (see Kotseruba & Tsotsos 2017, Lázaro-Gredilla et al. 2019, for examples).

## 3.2 Mined Methods

On viewing the graph of Figure 4A, and remembering there are hundreds of these collected, one might wonder whether there are sub-graphs that are repeated, and that perhaps they form standard chunks[5] of actions, in other words, the *methods* of our CP formulation. In fact, this appears to be the case. A total of 50 such methods represented by directed sub-graphs have been found, each with different usage frequency depending on target complexity, differences in target pose and subject starting position. Two examples are shown in Figure 6. Each of these combine 2 or more of the 50 methods into a kind of Bayes Net representation because they contain similar elements only with a different frequency of observation. Decision points are shown as little circles in the graph with relative frequency usage labeled on the outgoing arcs. Instances of each are circled in Figure 4A as examples (note that the top one is also part of the expanded view in Figure 5 so details can be seen).

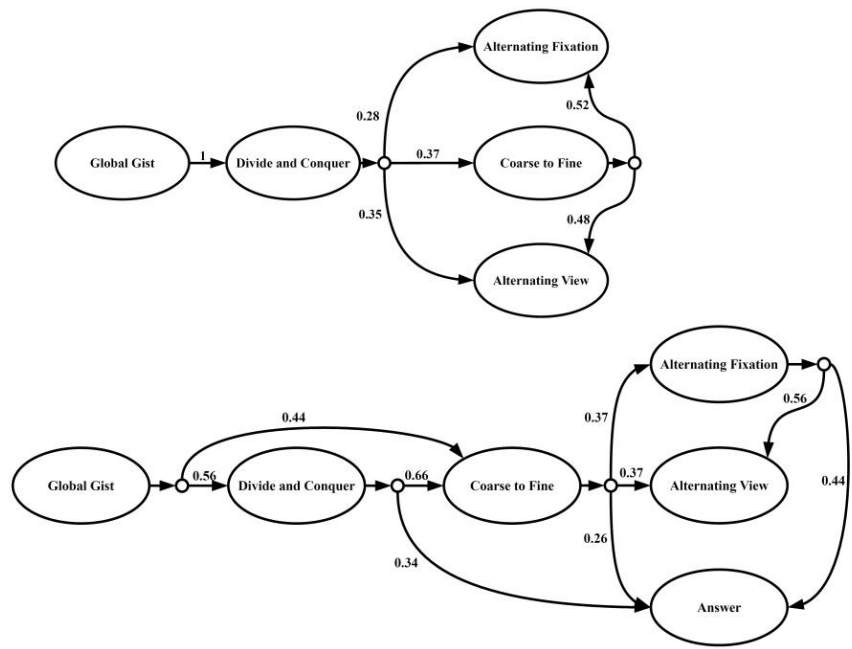

*Figure 6.* Two of the methods found in subject trials. The labels on the arcs between nodes give the frequency of occurrence for that arc. The small circles are choice points. Top: This sub-graph of actions appears in all trials that involved *easy* objects, whether *same* or *different*. Bottom: This sub-graph of actions was found in 99.7% of all trials that involved 2 *same* objects. It covers 4 of the 50 methods found (4 different pathways through that graph). The bottom right node is labelled 'Answer', and this means that this sub-graph was found as the last sub-graph in the sequence for these cases. Although different in content, these are like the CPs described in Tsotsos & Kruijne (2014) or in Kotseruba & Tsotsos (2017).

In one sense, each seems self-evident; yet each requires a much more detailed explanation which is beyond the scope of this paper (see Solbach & Tsotsos forthcoming). For example, *Global Gist* is an abstract concept that, as described above, tracks and controls how much of the virtual viewing sphere around each object has been explored. *Coarse-to-Fine* is similarly abstract and is characterized as increasing the amount and duration of fixations of parts of the object. Each involves complex internal actions on its own. In other words, it is expected that these methods will have an internal structure, which itself may be a composition of other methods. At their most primitive, some methods may be innate, but most would be learned.

---

[5] The idea of *chunking* in cognitive science goes back to at least Johnson (1970). He used the concept to group together elements of a behavior sequence and linked these to memory and their coding. Here we use the term *method* in keeping with the Cognitive Programs terminology introduced in Tsotsos & Kruijne (2014).

Figure 7 illustrates what might be considered such an internal structure for *Alternating Fixation*, defined in Section 2.5, and a component of both CP's in Figure 6. This is extracted from an actual trial and shows gaze moving from one object to the other, seemingly examining a small portion of the physical structure of each object, presumably to determine if that particular section was the same or different. To be sure, this characterization is preliminary and a great deal of detailed analysis of this data remains.

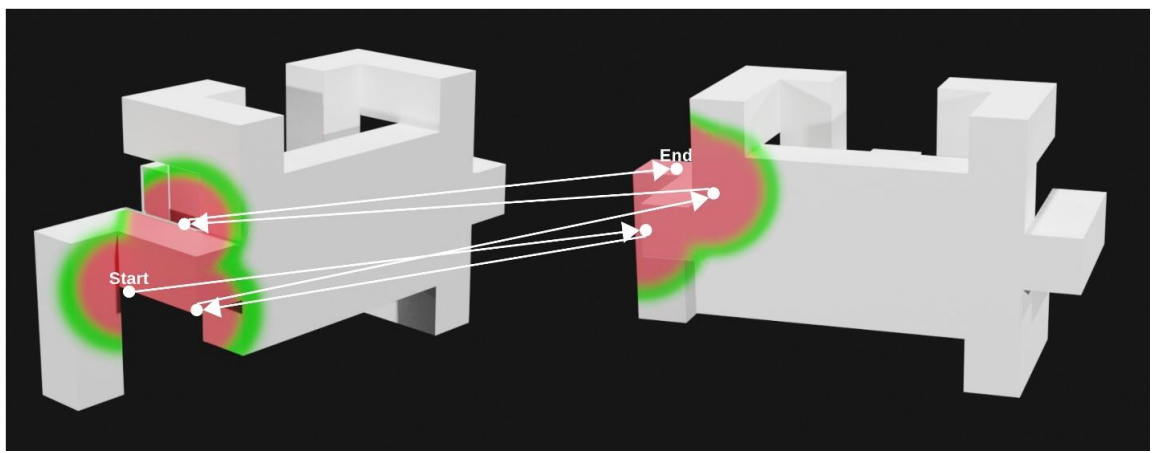

*Figure 7.* Here, we show a visualization of a fixation pattern that we identify as *Alternating Fixation* (See 2.5 B 5.). Both objects are displayed in the orientation of their observation. The corresponding fixations are highlighted with red circles and a green border. Arrows point to and originate at center of gaze. Further, the starting fixation is provided (annotated with 'Start'), and the subsequent fixation is connected with an arrow. The alternating fixation ends at the fixation marked with 'End.' The two objects are of complexity *medium*, they are the *same* object, presented at 90° orientational difference.

As described in the Introduction, we consider the visual system as a general-purpose processor, tuned to the task and input of the moment. It might be that you have learned thousands of such CPs, and you have them stored in memory, quickly deploying the right one(s) at the right time, dynamically parameterized for the task. This might suggest the need for a controller of some sort, and although there is controversy about such a need, as Tsotsos et al. (2021) describe, both theoretical and experimental arguments in that paper point to the clear importance of a controller. Some computational structure must be present to accept task instructions and translate them into executable CPs for solving the task. The timings of these actions need synchronization and coordination. How exactly this might occur is too much for the scope of this paper. However, the feasibility of the idea has been documented not only in the earlier VR work but also in our more current CP work (Abid 2018, Kunić 2017, Kotseruba & Tsotsos 2017, Tsotsos & Kruijne 2014). Abid used CPs to represent and test human behavior for several attentional psychophysical experiments. Kunić developed a compiler that would accept Imperative English and provide task specification for a CP while in Kotseruba & Tsotsos, two video games were encoded and tested to human top rank performance. The current results lend further support to CPs' potential for complex real-world tasks.

## 4. Conclusions

This paper outlines our first steps towards an understanding of how agents (human or artificial) might solve complex 3D visuospatial tasks as active observers. The original motivation was to flesh out how a rich set of attentional mechanisms might participate in our full cognitive architecture, STAR. We discovered that there was no human experimental data available to inform our work. Our primary goal then became to explore and identify the characteristics of human behavior in active tasks that could then inform the development of cognitive systems that perform the similar tasks. To this end, we developed a unique experimental facility and conducted targeted experiments to extend the classic 2D same-different task to 3D real objects observed by active subjects. Obvious problem instances were simply solved, but as we increased the difficulty, the apparent simplicity of the task masked the surprising complexity of human strategies deployed.

Here, we presented a few of the results, and data collection and analysis continues. Among the most striking of the results are the following:
• People are very good at this task even for difficult cases. The range of response times from simplest to most difficult cases ranged from 4 - 298 sec. and accuracy from 80% to 100%.
• There is a great deal of data acquisition occurring during all trials with the range of eye movements (and thus separate fixations and separate images processed) from 6 to 800 fixations.
• Fixation sequences seemed purposeful both as reported by subjects and by examining the sequences. We were able to identify 50 patterns of actions - *Cognitive Program methods* - that were repeated in various combinations in all trials.
• The CP methods as described must be considered as simplified. That is, each of the nodes in the sequence represents yet another sequence of actions including less abstract computations such as image analysis, attention, use of working memory, decision-making, planning and re-planning, etc.
• No statistical change has been observed (yet) in accuracy, number of fixations and overall head movement with increasing trials for individual subjects. This implies that each problem instance leads to a unique deployment of solution strategies, not a learned one.

This particular task seems an excellent testbed for testing systems that purport intelligent behavior. The Turing Test, just as an example, does not test active observation in the ways our task requires. There is no claim that this should replace it, of course; nevertheless, our data does point to a dimension of intelligence - the ability to decide how, why, when, what and where to sense the environment to best complete a task - that has not been well studied.

We have proposed that Cognitive Programs provide a flexible, dynamic composition of potential solutions. We intend to also experiment with 3D 'spatial relations', 3D 'visual search', and to add shadowing to all the tasks, within the PESAO facility. The goal is not to solve each separately but rather to discover the common elements of a generic visual problem-solving strategy.

Finally, the attentional mechanisms apparent in these experiments show more breadth than those in common computational use and agree with the mechanisms of ST. Specifically:

• Priming - If looking for specific structure seen on one object in the other, then the system is primed for it. In ST priming manifests as a suppression of competing features (e.g., Rosenfeld et al. 2018).
• Fixation change - Internal attention shifts precede eye movement and attention-based head movement. As shown in Tsotsos et al. (2016) and Wloka et al. (2018), this computation requires a combination of spatially central object attention, feature saliency-based peripheral attention, fixation history and task priming.
• Surround suppression - Comparison of structures requires an inspection of differences in shape, size and position, going beyond simple classification, and as Yoo et al. (2019) showed, this requires top-down surround suppression and localization.
• Selection - The strategy graphs demonstrate that choice between alternatives is required often. Selection mechanisms are a key part of attention and play a role at sensory, cognitive and behavioral levels, as ST includes (Tsotsos 2011).

It seems that the original motivation for STAR, to test if the Selective Tuning (ST) model of attention provides all the attentional support a behaving brain (agent) would require, gains good support based on the experiments and analysis presented here. Knowing what a solution to the 3D visuospatial task Same-Different looks like, the effort to build a cognitive system (such as STAR) to perform similarly has a much better grounding. The reality of active human behavior will likely reveal many surprises to come, and with that, many challenges for how artificial agents may be developed with the same abilities.

**Acknowledgements**

This research was supported by several sources for which the authors are grateful: Air Force Office of Scientific Research (FA9550-18-1-0054), the Canada Research Chairs Program (950-231659), and the Natural Sciences and Engineering Research Council of Canada (RGPIN-2016-05352).